\def\year{2022}\relax
\documentclass[letterpaper]{article} 
\usepackage{aaai22}  
\usepackage{times}  
\usepackage{helvet}  
\usepackage{courier}  
\usepackage[hyphens]{url}  
\usepackage{graphicx} 
\usepackage{natbib}  
\usepackage{caption} 
\DeclareCaptionStyle{ruled}{labelfont=normalfont,labelsep=colon,strut=off} 
\usepackage{algorithm}
\usepackage{algorithmic}

\usepackage{newfloat}
\usepackage{listings}
\floatstyle{ruled}
\newfloat{listing}{tb}{lst}{}
\floatname{listing}{Listing}
\usepackage{researchpack}
\usepackage{xcolor}
\usepackage{xspace}
\usepackage{booktabs}

\newcommand{\defeq}{\ensuremath{:=}}

\newcommand{\expl}{\ensuremath{\calE}\xspace}
\newcommand{\dataset}{\ensuremath{D}\xspace}
\newcommand{\memory}{\ensuremath{M}\xspace}
\newcommand{\act}{\ensuremath{\mathrm{act}}\xspace}
\newcommand{\attr}{\ensuremath{\mathrm{attr}}\xspace}
\newcommand{\aggr}{\ensuremath{\mathrm{aggr}}\xspace}
\newcommand{\lattr}{\ensuremath{\ell_{\attr}}}
\newcommand{\laggr}{\ensuremath{\ell_{\aggr}}}

\title{Toward a Unified Framework for Debugging Concept-based Models}
\author {
    Andrea Bontempelli,\textsuperscript{\rm 1}
    Fausto Giunchiglia,\textsuperscript{\rm 1,2}
    Andrea Passerini,\textsuperscript{\rm 1}
    Stefano Teso,\textsuperscript{\rm 1}
}
\affiliations {
    \textsuperscript{\rm 1} University of Trento,  Italy\\
    \textsuperscript{\rm 2} Jilin University, China\\
    name.surname@unitn.it
}

\begin{document}
\maketitle

\begin{abstract}
    In this paper, we tackle interactive debugging of ``gray-box'' concept-based models (CBMs).  These models learn task-relevant concepts appearing in the inputs and then compute a prediction by aggregating the concept activations.
    Our work stems from the observation that in CBMs \emph{both} the concepts \emph{and} the aggregation function can be affected by different kinds of bugs, and that fixing these bugs requires different kinds of corrective supervision.
    To this end, we introduce a simple schema for human supervisors to identify and prioritize bugs in both components, and discuss solution strategies and open problems.
    We also introduce a novel loss function for debugging the aggregation step that generalizes existing strategies for aligning black-box models to CBMs by making them robust to how the concepts change during training.
\end{abstract}

\section{Introduction}

A central tenet of eXplainable AI (XAI) is that explanations of a model's predictions naturally uncover bugs and biases affecting the model~\cite{lapuschkin2019unmasking,schramowski2020making}.
Post-hoc explanations of black-box models, however, can be unfaithful and ambiguous~\cite{dombrowski2019explanations,teso2019toward,lakkaraju2020fool,sixt2020explanations}, and the extraction process can be computationally challenging~\cite{van2021tractability}.

Concept-based models (CBMs) are designed to make the extraction step as straightforward as possible while retaining the performance of more opaque alternatives~\cite{rudin2019stop}.
To this end, CBMs learn a set of high-level, interpretable concepts capturing task-salient properties of the inputs, and then obtain predictions by aggregating the concept activations in a (typically) understandable manner~\cite{alvarez2018towards,losch2019interpretability,koh2020concept,chen2019looks,hase2019interpretable,rymarczyk2020protopshare,nauta2021neural,lage2020learning}.
A key feature of CBMs is that they explain their own predictions by supplying faithful concept-level attribution maps, which encompass both the concepts and the aggregation weights, thus facilitating the identification of bugs affecting the model.

Work on interactive troubleshooting of CBMs is, however, sparse and \emph{ad hoc}:  some approaches assume the concepts to be given and high-quality and focus on correcting the aggregation step~\cite{teso2019toward,stammer2021right}, others address issues with the learned concepts while ignoring how they are aggregated~\cite{barnett2021iaia,lage2020learning}.  Fixing only one set of bugs is however insufficient.

In this paper, we outline a unified framework for debugging CBMs.
Our framework stems from the simple observation that the quality of a CBM hinges on \emph{both} the concepts vocabulary \emph{and} on how the concepts are aggregated, and that both elements are conveniently captured by the CBM's explanations.
This immediately suggests a human-in-the-loop debugging strategy based on providing \emph{supervision on the model's explanations} that is composed of three different phases, namely (i) evaluating concept quality, (ii) correcting the aggregation weights, and (iii) correcting the concepts themselves.
This novel yet intuitive setup allows us to identify limitations in existing works and highlight possible ways to overcome them.

As a first step toward implementing this framework, we introduce a new loss function on the aggregation weights that generalizes approaches for aligning local explanations~\cite{ross2017right,lertvittayakumjorn2020find} to be robust to changes in the underlying concepts during training.
We then outline how the same strategy can be applied recursively to align the concepts themselves by correcting both their predictions and explanations.

\paragraph{Contributions.}  Summarizing, we:
\begin{enumerate}

    \item Introduce a unified framework for debugging CBMs that explicitly distinguishes between and addresses bugs affecting how the concepts are defined and how they are aggregated.

    \item Illustrate how to incorporate corrective feedback into the aggregation step in a manner that is invariant to changes to the learned concepts.

    \item Discuss how to align the concepts by fixing the reasons behind their activations, opening the door to explanation-based debugging of the concepts themselves.

\end{enumerate}

\section{Concept-based Models}

We are concerned with learning a high-quality classifier $f: \vx \mapsto y$ that maps instances $\vx \in \bbR^d$ into labels $y \in [v] \defeq \{1, \ldots, v\}$.
In particular, we focus on \emph{concept-based models} (CBMs) that fit the following two-level structure.

At the lower level, the model extracts a vector of \emph{concept activations}
$
    \vc(\vx) = (c_1(\vx), \ldots, c_k(\vx)) \in \bbR^k
$
from the raw input $\vx$.
The concepts $\{ c_j \}$ are usually learned from data so to provide strong indication of specific classes~\cite{chen2019looks}.  For instance, in order to discriminate between images of cars and plants, the model might learn concepts that identify wheels and leaves.
The concepts themselves are completely black-box.

At the upper level, the model \emph{aggregates} the concept activations into per-class scores, typically in a simulatable~\cite{lipton2018mythos} manner.  This is most often implemented by taking a linear combination of the activations:
\[
    \textstyle
    s_y(\vx) \defeq \inner{\vw^{(y)}(\vx)}{\vc(\vx)} = \sum_j w_j^{(y)}(\vx) \cdot c_j(\vx)
    \label{eq:scores}
\]
where $\vw^{(y)}(\vx) \in \bbR^k$ is the weight vector associated to class $y$.
Class probabilities are then obtained by passing the scores through a softmax activation, that is,
$
    P(y \ |\ \vx) \defeq \mathrm{softmax}( \vs(\vx) )_y
$.

CBMs are learned by minimizing an empirical loss $1 / |\dataset| \cdot \sum_i \ell(f, (\vx_i, y_i))$, where $i$ runs over the training set \dataset, as customary.
The loss function $\ell(f, (\vx, y))$ itself combines a standard loss for classification, like the cross-entropy loss, together with additional regularization terms that encourage the learned concepts to be understandable to (sufficiently expert) human stakeholders.  Commonly used criteria include similarity to concrete examples~\cite{alvarez2018towards,chen2019looks}, disentanglement~\cite{alvarez2018towards}, and boundedness~\cite{koh2020concept}.

With CBMs, it is straightforward to extract local explanations that capture how different concepts contribute to a decision $(\vx, y)$.  These explanations take the form:
\[
    \textstyle
    \expl(\vx, y) \defeq \{(w_j^{(y)}(\vx),\, c_j(\vx)) : j \in [k]\}
    \label{eq:gbm-explanation}
\]
Notice that \emph{both the concepts and the aggregation weights are integral to the explanation}: the concepts $\{c_j\}$ establish a vocabulary that enables communication with stakeholders, while the weights $\{w_j(\vx)\}$ convey the relative importance of different concepts.
Crucially, the score of class $y$ is independent from $\vx$ given the explanation $\expl(\vx, y)$, ensuring that the latter is faithful to the model's decision process.

\begin{table*}[t]
    \centering
    \begin{small}
    \begin{tabular}{ccccc}
        \toprule
        \textsc{Method}
            & \textsc{Concepts} $\vc(\vx)$
            & \textsc{Aggregator} $f(\vc)$
            & \textsc{Extra Annot.}
            & \textsc{Training}
        \\
        \midrule
        SENN~\cite{alvarez2018towards}
            & Autoencoder
            & Linear Comb.
            & --
            & End-to-end
        \\
        ProtoPNet~\cite{chen2019looks}
            & Conv. Filters
            & Linear Comb.
            & --
            & Multistep
        \\
        IAIA-BL~\cite{barnett2021iaia}
            & Conv. Filters
            & Linear Comb.
            & Concept Attr.
            & Multistep
        \\
        CBM~\cite{koh2020concept}
            & Arbitrary
            & Arbitrary
            & Concept Labels
            & End-to-end
        \\
        \bottomrule
    \end{tabular}
    \end{small}
    \caption{Comparison between concept-based CBMs considered in this work.}
    \label{tab:gbms}
\end{table*}

\subsection{Implementations}

Next, we introduce some well-known CBMs that match the above template.  A summary can be found in Table~\ref{tab:gbms}.  Additional models are briefly discussed in the Related Work.

\medskip \noindent \textbf{Self-Explainable Neural Networks} (SENNs), introduced by \citet{alvarez2018towards}, generalize linear models to representation learning.
SENNs instantiate the above two-level template as follows.
The weight functions $\vw^{(y)}(\vx)$, $y \in [v]$, are implemented using a neural network for multivariate regression.  Importantly, the weights are regularized so to vary slowly across inputs.  This is achieved by penalizing the model for deviating from its first-order (i.e., linear) Taylor decomposition.
The concepts $\vc(\vx)$ are either given and fixed or learned using a sparsity-regularized autoencoder, and once learned they are conveyed to users by presenting a handful of concrete training examples on which the concepts maximally activate.
All components of SENNs are training jointly in an end-to-end fashion.

\medskip \noindent \textbf{Part Prototype Networks} (ProtoPNets)~\cite{chen2019looks} ground the two-level template to image classification as follows.
The weights $\vw^{(y)}$ are constant with respect to the input $\vx$, while the concepts $\vc(\vx)$ indicate the presence of ``part-prototypes'', that is, prototypes that capture specific parts of images appearing in the training set.
Specifically, the part-prototypes are implemented as (parts of) points in the embedding space of a (pre-trained) convolutional neural network.  Each part-prototype $\vp$ is encoded as a $1 \times 1 \times q$ parameter vector and activates based on the contents of a corresponding rectangular receptive field in input space.  Each class $y \in [v]$ is associated to its own $\lfloor\frac{k}{v}\rfloor$ part-prototypes, which are learned so as to maximally activate on (parts of) training images of the associated class and not to activate on those of the other classes.
The concepts and the weights are learned sequentially.

\medskip \noindent \textbf{IAIA-BL}~\cite{barnett2021iaia} specializes ProtoPNets to image-based medical diagnosis.  In particular, in addition to label annotations, IAIA-BL accepts per-example attribute relevance information (e.g., annotations of symptomatic regions in X-ray images) and penalizes part-prototypes that activate outside the relevant areas.

\medskip \noindent \textbf{Concept Bottleneck Models} (CBNMs)~\cite{koh2020concept,losch2019interpretability} are regular feed-forward neural networks in which the neurons in a given layer are trained to align with a known vocabulary of concepts $\vc(\vx)$.  This is achieved by supplying the CBNM with concept-level annotations.
The aggregation step is not particularly restrained:  the concept neurons are either using a single dense layer, in which case the weights $\vw^{(y)}$ are constant with respect to $\vx$, or through a sequence of layers, in which case the weights $\vw^{(y)}(\vx)$ are \emph{not} constant and must be inferred using post-hoc techniques (e.g., input gradients).

\section{Existing Strategies for Debugging CBMs}

Existing literature on debugging CBMs can be split into two groups.
Approaches in the first group are concerned with bugs in the aggregation weights $\{\vw^{(y)}(\vx)\}_y$.  These can occur when the data fools the model into assigning non-zero weight to concepts that correlate with -- but are not causal for -- the label.  A prototypical example are class-specific watermarks in image classification~\cite{lapuschkin2019unmasking}.
\citet{teso2019toward} addresses this issue by leveraging explanatory interactive learning (XIL)~\cite{teso2019explanatory,schramowski2020making}.  As in active learning, in XIL the machine obtains labels by querying a human annotator, but in addition it presents predictions for its queries and local explanations for its predictions.  The user then provides corrective feedback on the explanations, for instance by indicating those parts of an image that are irrelevant to the class label but that the machine relies on for its prediction.  The model is then penalized whenever its weights do not align with the user's corrections.  This setup assumes that the concepts $\vc$ are fixed rather than learned from data.
\citet{stammer2021right} extend XIL to neuro-symbolic models and structured attention, but also assume the concepts to be fixed.

Conversely, approaches in the second group are only concerned with how the concepts are defined:  concepts learned from data, even if discriminative and interpretable, may be misaligned with (the stakeholders's understanding of) the prediction task and may thus fail to generalize properly.
CBNMs work around this issue by leveraging concept-level label supervision~\cite{koh2020concept}, however this does not ensure that the concepts themselves are ``right for the right reasons''~\cite{ross2017right}.
As a matter of fact, just like all other neural nets~\cite{szegedy2013intriguing}, part-prototypes learned by ProtoPNets can pick up and exploit uninterpretable features of the input~\cite{hoffmann2021looks}.
\citet{hoffmann2021looks} and \citet{nauta2020looks} further argue that it may be difficult for users to understand what a learned prototype (concept) represents, unless the model explains \emph{why} the concept activates.  To this end, \citet{nauta2020looks} propose a perturbation-based technique -- analogous to LIME~\cite{ribeiro2016should} -- for explaining why a particular prototype activates on a certain region of an image, which however does nothing to \emph{fix} the bugs that it highlights.
Finally, \citet{barnett2021iaia} introduce a loss term for ProtoPNets that penalizes concepts that activate on regions annotated as irrelevant by a domain expert.

\section{A Unified Framework for Debugging CBMs}

We propose a new unified framework -- based on the right for the right reasons (RRR) principle~\cite{ross2017right} -- that, in contrast with existing solutions, is designed for the more realistic case of CBMs affected by multiple, different bugs.

The RRR principle is straightforward:  the goal is to ensure that the model outputs accurate \emph{predictions} that are justified by high-quality \emph{explanations}.  In order to make this concrete, we need to define what we mean for an explanation $\expl(\vx, y)$ to be ``high-quality''.  Intuitively, for CBMs explanation quality depends on the quality of the learned concepts $\vc$ and aggregation weights $\vw(\vx)$.  We unpack this intuition as follows:

\begin{defn}
A set of concepts $\vc$ is high-quality for a decision $(\vx, y)$ if:
\begin{itemize}

    \item[\textbf{C1.}] The set of concepts is sufficient to fully determine the \emph{ground-truth} label $y^*$ from $\vx$.\footnote{In case the ground-truth label is ill-defined, it should be replaced with the Bayes optimal label.}
    
    \item[\textbf{C2.}] The various concepts are (approximately) independent from each other.

    \item[\textbf{C3.}] Each concept is semantically meaningful.

    \item[\textbf{C4.}] Each concept is easy to interpret (e.g., simple enough).

\end{itemize}
Given high-quality concepts $\vc$, a set of weights $\vw^{(y)}(\vx)$ is high quality for a decision $(\vx, y)$ if:
\begin{itemize}

    \item [\textbf{W1.}] It ranks all concepts in $\vc(\vx)$ compatibly with their relevance for the prediction task.
    
    \item [\textbf{W2.}] It associates (near-)zero weight to concepts in $\vc(\vx)$ that are task-irrelevant.
    
\end{itemize}
An explanation $\expl(\vx, y)$ is high-quality for a decision $(\vx, y)$ if both $\vc$ and $\vw^{(y)}(\vx)$ are.
\end{defn}

\noindent
It is worth discussing the various points in detail.
Requirement C1 ensures that the concepts are task-relevant and jointly sufficient to solve the prediction task.  This requirement is imposed by the learning problem itself.  Notice that, since in all CBMs the prediction $f(\vx)$ is independent from the input $\vx$ given the explanation $\expl(\vx, f(\vx))$, C1 is necessary for the model to achieve good generalization.
Although some works stress the need for the concept vocabulary to be complete~\cite{yeh2020completeness,bahadori2021debiasing}, we argue that -- when it comes to local explanations -- sufficiency is more relevant, although it is clear that if the concepts $\vc$ are sufficient for all instances $\vx$, then they do form a complete vocabulary.

Requirements C2, C3, and C4, on the other hand, are concerned with whether the concept set can be used for \emph{communicating} with human stakeholders, and lie at the core of CBMs and human-in-the-loop debugging.
Notice also that requirements C1--C4 are mutually independent: not all task-relevant concepts are understandable (indeed, this is often not the case), not all semantically meaningful concepts are easy to interpret, \emph{etc.}

\subsection{Debugging CBMs in Three Steps}

Assume to be given a decision $(\vx, y)$ and an explanation $\expl(\vx, y)$ that is \emph{not} high quality.  What bugs should the user prioritize?  We propose a simple three-step procedure:
\begin{description}

    \item[Step 1] \emph{Evaluating concept quality:} Determine if $\expl(\vx)$ contains a high-quality subset $\vc' \subseteq \vc$ that is \emph{sufficient} to produce a correct prediction $y^*$ for the target instance $\vx$.

    \item[Step 2] \emph{Correcting the aggregation weights:}  If so, then it is enough to fix how the model combines the available concepts by supplying corrective supervision for the aggregation weights $\vw$.

    \item[Step 3] \emph{Correcting the learned concepts:} Otherwise, it is necessary to create a high-quality subset $\vc'$ by supplying appropriate supervision on the concepts $\vc$ themselves.

\end{description}
In the following we discuss how to implement these steps.

\subsection{Step 1: Assessing concept quality}

In this step, one must determine whether there exists a subset $\vc' \subseteq \vc$ that is high-quality.  This necessarily involves conveying the learned concepts $\vc$ to a human expert in detail sufficient to determine whether they are ``good enough'' for computing the ground-truth label $y^*$ from $\vx$.

The most straightforward solution, adopted for instance by SENNs~\cite{alvarez2018towards} and ProtoPNets~\cite{chen2019looks}, is to present a set of instances that are most \emph{representative} of each concept.  In particular, ProtoPNets identify parts of training instances $\vx$ that maximally activate each $c_j \in \vc$.
A more refined way to characterize a concept is to explain \emph{why} it activates for certain instances, possibly the prototypes themselves~\cite{nauta2020looks,hoffmann2021looks}.  Such explanations can be obtained by extracting an attribution map $\attr(c_j, \vx)$ that identifies those inputs -- either raw inputs $\vx$ or higher-level features of $\vx$ like color, shape, or texture -- that maximally contribute to the concept's activations.  Since concepts are black-box, this map must be acquired using post-hoc techniques, like input gradients~\cite{baehrens2010explain,sundararajan2017axiomatic} or LIME~\cite{ribeiro2016should}.  Explanations are especially useful to prevent stakeholders from confusing one concept for another.  For instance, in a shape recognition task, a part-prototype that activates on a yellow square might do so because of the shape, of the color, or both.  Without a proper explanation, the user cannot tell these concepts apart.  This is fundamental for preventing users from wrongly trusting ill-behaved concepts~\cite{teso2019explanatory,rudin2019stop}.

Although assessing concept quality does \emph{not} require to communicate the full semantics of a learned concept to the human counterpart, acquiring feedback on the aggregation weights \emph{does}.
Doing so is non-trivial~\cite{zhang2019dissonance}, as concepts learned by CBMs are not always interpretable~\cite{nauta2020looks,hoffmann2021looks}.  However, we stress that uninterpretable concepts are not high-quality, and therefore they must be dealt with in Step 3, i.e., they must be improved by supplying appropriate concept-level supervision.

\subsection{Step 2: Fixing how the concepts are used}

In this second step, the goal is to fix up how the concepts are aggregated.  This case is reminiscent of debugging black-box models.  Existing approaches for this problem acquire a (possibly partial~\cite{teso2019explanatory,teso2019toward}) ground-truth attribution map $\vm \in \{0, 1\}^d$ that specifies what input attributes are (ir)relevant for the target prediction, and then penalize the model for allocating non-zero relevance to them.

More specifically, let $f$ be a \emph{black-box} model and $\attr(f, (\vx, y)) \in \bbR^d$ be an attribution mechanism that assigns numerical responsibility for the decision $(\vx, y)$ to each input $x_i$, for $i = 1, \ldots, d$, for instance integrated gradients~\cite{sundararajan2017axiomatic}.
The model's explanations are corrected by introducing a loss of the form~\cite{ross2017right,schramowski2020making,shao2021right}:\footnote{The loss can be adapted to also encourage the model to rely on concepts that are deemed relevant, for instance by penalizing the model whenever the concept's weight is too close to zero.}
\[
    \textstyle
    \lattr(f, (\vx, y), \vm) \defeq \sum_{i \in [d]} (1 - m_i) \cdot \attr_{i}(f, (\vx, y))^2
    \label{eq:attr-loss}
\]

Now, consider a CBM $f$ and a decision $(\vx, y)$ obtained by aggregating \emph{high-quality} concepts $\vc$ using \emph{low-quality} weights $\vw(\vx)$.  It is easy to see how Eq.~\ref{eq:attr-loss} could help with aligning the aggregation weights.  However, simply replacing $\attr_i$ with the weights $\vw$ does not work.  The reason is that in CBMs the concepts \emph{change} during training, and so do the semantics of their associated weights.  Hence, feedback of the form ``don't use the $j$-th concept'' becomes obsolete (and misleading) whenever the $j$-th concept changes.

Another major problems with Eq.~\ref{eq:attr-loss} is that, by penalizing concepts by their index, it does not prevent the model from re-learning a forbidden concept under a different index.
Consider a toy image classification task in which the positive class consists of images that contains a green circle (50\% of the images) or a pink triangle (the other 50\%), but the data is confounded such that 100\% of the positive training images also contain a yellow square.  A ProtoPNet with a budget of two prototypes per class is encouraged to rely on the counfounder to achieve 100\% accuracy on the training set.  If the model is penalized with Eq.~\ref{eq:attr-loss} for using the confounder as part-prototype $1$, it still has plenty of room to learn the confounder using the \emph{other} prototype of the positive class.

Roughly speaking, this shows that CBMs are too flexible for regular alignment losses and can therefore easily work around them.  Our solution to this problem is presented in the next Section.

\subsection{Step 3: Fixing how the concepts are defined}

Finally, consider a decision $(\vx, y)$ that depends on a \emph{low-quality} set of concepts $\vc$.  In this case, the goal is to ensure that at least some of those concepts quickly become useful by supplying additional supervision.

The order in which the various concepts should be aligned is left to the user.  This is reasonable under the assumption that she is a domain expert and that those concepts that are worth fixing are easy to distinguish from those that are not.
An alternative is to debug the concepts sequentially based on their overall impact on the model's behavior, for instance sorting them by decreasing relevance $|w_j^{(y)}(\vx)|$.  More refined strategies will be considered in future work.

Now, let $c_j$ be a low-quality concept.  Our key insight is that there is no substantial difference between the models' output $f(\vx)$ and a specific concept $c_j(\vx)$ appearing in it.\footnote{A similar point was brought up in~\cite{stammer2021right}.}  This means that work on understanding and correcting predictions of black-box models can be applied for understanding and correcting concepts in CBMs -- with some differences, see below.  For instance, the work of~\cite{nauta2020looks} can be viewed as a concept-level version of LIME~\cite{ribeiro2016should}.  To the best of our knowledge, this useful analogy has never been pointed out before.

A direct consequence is that concepts, just like predictions, can be aligned by providing labels for them, as is done by CBMs~\cite{koh2020concept}.  One issue with this strategy is that, if these annotations are biased, the learned concept may end up relying on confounders~\cite{lapuschkin2019unmasking}.
A more robust alternative, that builds on approaches for correcting the model's explanations~\cite{ross2017right,teso2019explanatory}, is to also align the concepts' \emph{explanations}.  This involves extracting an explanation $\attr(c_j, \vx)$ that uncovers the reasons behind the concept's activations in terms of either inputs, as done in~\cite{barnett2021iaia}, or higher-level features, and supplying corrective feedback for them.
One caveat is that these strategies are only feasible when the semantics of $c_j$ are already quite clear to the human annotator (e.g., the concept clearly captures ``leaves'' or ``wheels'', rather than seemingly random but class-discriminative blobs, as can occur during training).  If this is not the case, then the only option is to instruct the model to avoid using it by using \laggr or analogous penalties.

\paragraph{On-demand ``conceptification''.}  Since concepts in CBMs are black-box, explanations for them must be extracted using post-hoc attribution techniques, which -- as we argued in the introduction -- is less than optimal.  A better alternative is to model the concepts themselves as CBMs, making cheap and faithful explanations \emph{for the concepts} readily available.  We call the idea of replacing a black-box model with an equivalent CBM ``\emph{conceptification}''.  Said CBM can be obtained using, for instance, model distillation~\cite{gou2021knowledge}, and it would persist over time.  A nice benefit of conceptification is that all debugging techniques discussed so far would immediately become available for the newly introduced concepts too.

Conceptification makes the most sense in applications where concepts can be computed from simpler, lower-level concepts (like shape, color, and texture~\cite{nauta2020looks}) or where concepts exhibit a hierarchical, parts-of-parts structure~\cite{fidler2007towards}.   Notably, conceptification needs not be applied to all concepts unconditionally.  One interesting direction of research is to let the user decide what concepts should be conceptified -- for instance those that need to be corrected more often, or for which supervision can be easily provided.

\subsection{The Debugging Loop}

Summarizing, interactive debugging of CBMs involves repeatedly choosing an instance $(\vx, y)$ and then proceeding as in steps 1--3, acquiring corrective supervision on weights and concepts along the way.
The target decisions can be either selected by the machine, as in explanatory active learning~\cite{teso2019explanatory}, or selected by a human stakeholder, perhaps aided by the machine~\cite{popordanoska2020machine}.
Since in CBMs concepts and weights are learned jointly, the model is retrained to permit supervision to flow to both concepts and weights, further aligning them.  This retraining stage proceeds until the model becomes stable enough in terms of, e.g., prediction or explanation accuracy on the training set or on a separate validation set.
At this point, the annotator can either proceed with another debugging session or stop if the model looks good enough.

This form of debugging is quite powerful, as it allows stakeholders to identify and correct a variety of different bugs, but it is not universal: some bugs and biases that cannot be uncovered using local explanations~\cite{popordanoska2020machine}.  Other bugs -- like those due to bad choice of hyper-parameters, insufficient model capacity, and failure to converge -- are not fixable with any form of supervision.  Dealing with these issues, however, is beyond the scope of this paper and left to future work.

\section{A Robust Aggregation Loss}
\label{sec:aggr-loss}

Simply discouraging the model from using incorrectly learned concepts does not prevent it from learning them again with minimal changes.  To address this problem, we propose to penalize the model for associating non-zero weight to concepts that are \emph{similar} to those concepts ${c}$ indicated as irrelevant to the target decision during the debugging session.  We assume these concepts to have been collected in a memory $\memory$.  The resulting loss is:
\[
    \laggr(f, (\vx, y), \memory) \defeq \sum_{\substack{c \in \memory(\vx, y) \\ j \in [k]}} w_j^{(y)}(\vx)^2 \cdot \kappa({c}, c_j)
    \label{eq:aggr-loss}
\]
The sum iterates over old concepts ${c}$ irrelevant to $(\vx, y)$, denoted $\memory(\vx, y)$, and over the current concepts $\vc$, while $0 \le \kappa(c,c') \le 1$ measures the similarity between concepts.

Notice how Eq.~\ref{eq:aggr-loss}, being independent of the concept's order, prevents the model from re-learning forbidden concepts.  Moreover, it automatically deals with redundant concepts, which are sometimes learned in practice~\cite{chen2019looks}, in which cases all copies of the same irrelevant concept are similarly penalized.

\paragraph{Measuring concept similarity.}  A principled way of doing so is to employ a kernel between functions.  One natural approach is to employ a product kernel~\cite{jebara2004probability}, for instance:
\[
    \kappa_{\act}(c, c') \defeq \int_{\bbR^d} (c(\vx) \cdot c'(\vx))^\rho p^*(\vx) d\vx
    \label{eq:act-kernel}
\]
Here, $p^*(\vx)$ is the ground-truth distribution over instances and $\rho > 0$ controls the smoothness of the kernel.  It is easy to show that $\kappa_{\act}$ is a valid kernel by rewriting it as an inner product $\inner{(q^*(\vx) c(\vx))^\rho}{\ (q^*(\vx) c'(\vx))^\rho}$, where we set $q^*(\vx) \defeq p^*(\vx)^{1 / (2\rho)}$.

This kernel measures how often two concepts co-activate on different inputs, which is somewhat limiting.  For instance, when discriminating between images of cars and plants, the concepts ``wheel'' and ``license plate'' are deemed similar because they co-activate on car images and not on plant images.  This suggests using a more fine-grained kernel that considers \emph{where} the concepts activate, for instance:
\[
    \kappa_{\attr}(c, c') \defeq \int_{\bbR^d} \inner{\attr(c, \vx)}{\,\attr(c', \vx)}^\rho p^*(\vx) d\vx
    \label{eq:attr-kernel}
\]
Since co-localization entails co-activation, this kernel specializes $\kappa_{\act}$, in the sense that $\kappa_{\attr}(c, c') \le \kappa_{\act}(c, c')$.  Unfortunately, $p^*$ is not known and the integrals in Eqs.~\ref{eq:act-kernel} and~\ref{eq:attr-kernel} are generally intractable, so in practice we replace the latter with a Monte Carlo approximation on the training data.
Notice that the activation and localization maps of irrelevant concepts in $\memory$ can be cached for future use, thus speeding up the computation.

Certain models may admit more efficient kernels between concepts.  For instance, in ProtoPNets one can measure concept similarity by directly comparing the convolutional filter representations $\vp$ of the learned concepts, e.g., $\inner{\vp}{\vp'}$.
Using this kernel gives the following aggregation loss:
\[
        \sum_{\substack{\vp \in \memory(\vx, y) \\ j \in [k]}} \inner{{\vp}}{\vp_j} \cdot w_j(\vx)^2
        = \biginner{\sum_{\vp \in \memory(\vx, y)} {\vp}}{\sum_{j \in [k]} w_j(\vx)^2 \cdot \vp_j}
\]
which dramatically simplifies the computation.  However, the intuitive meaning of inner products in parameter space is not entirely obvious, rendering this kernel harder to control.  A thorough analysis of this option is left to future work.

\paragraph{Beyond instance-level supervision.}  Eq.~\ref{eq:aggr-loss} essentially encourages the distribution encoded by the CBM not to depend on a known-irrelevant concept.
In previous work, relevance annotations are provided at the level of individual examples and, as such, specify an invariance that only holds for few similar examples~\cite{teso2019explanatory,schramowski2020making,lertvittayakumjorn2020find,shao2021right,stammer2021right}.

Users, however, may find it more natural to provide relevance information for entire classes or for individual concepts.  For instance, although the concept of ``snow'' is not relevant for a specific class ``wolf'', it may be useful for other classes, like ``winter''.   Similarly, some concepts -- for instance, artifacts in X-ray images due to the scanning process -- should never be used for making predictions for any class.  These more general forms of supervision encode invariances that applies to entire \emph{sets} of examples, and as such have radically different expressive power.  A nice property of Eq.~\ref{eq:aggr-loss} is that it can easily encode such supervision by letting the outer summation range over (or at least sample) sets of instances, i.e., all instances that belong to a certain class or all instances alltogether.

\section{Related Work}

Our work targets well-known concept-based CBMs, including SENNs~\cite{alvarez2018towards}, CBMs~\cite{koh2020concept,losch2019interpretability}, and ProtoPNets~\cite{li2018deep,chen2019looks,hase2019interpretable,rymarczyk2020protopshare,barnett2021iaia} and related approaches~\cite{nauta2021neural}, but it could be easily extended to similar models and techniques, for instance concept whitening~\cite{chen2020concept}.  The latter is reminiscent of CBMs, except that the neurons in the bottleneck layer are normalized and decorrelated and feedback on individual concepts is optional.  Of course, not all CBMs fit our template.  For instance,  \cite{alshedivat2020contextual} propose a Bayesian model that, although definitely gray-box, does not admit selecting a unique explanation for a given prediction.  More work is needed to capture these additional cases.

Our debugging approach is rooted in explanatory interactive learning, a set of techniques that acquire corrective from a user by displaying explanations (either local~\cite{teso2019explanatory,selvaraju2019taking,lertvittayakumjorn2020find,schramowski2020making,shao2021right} or global~\cite{popordanoska2020machine}) of the model's beliefs.
Recently, \cite{stammer2021right} designed a debugging approach for the specific case of attention-based neuro-symbolic models.
These approaches assume the concept vocabulary to be given rather than learned  and indeed are not robust to changes in the concepts, as illustrated above.  Our aggregation loss generalizes these ideas to the case of concept-based CBMs.
Other work on XIL has focused on example-based explanations~\cite{teso2021interactive,zylberajch2021hildif}, which we did not include in our framework, but that could provide an alternative device for controlling a CBM's reliance on, e.g., noisy examples.

A causal approach for debiasing CBMs has been proposed by~\cite{bahadori2021debiasing}.  This work is orthogonal to our contribution, and could be indeed integrated with it.  The strategy of \cite{lage2020learning} acquires concept-based \emph{white} box models that are better aligned with the user's mental model by interactively acquiring concept-attribute dependency information.  The machine does so by asking questions like ``does depression depend on lorazepam?'', acquiring more impactful dependencies first.  This approach is tailored for specific white-box models, but it could and should be extended to CBMs and integrated into our framework.

\section{Conclusion}

We proposed a unified framework for debugging concept-based CBMs that fixes bugs by acquiring corrective feedback from a human supervisor.  Our key insight is that bugs can affect both how the concepts are defined and how they are aggregated, and that both elements have to be high-quality for a CBM to be effective.  We proposed a three-step procedure to achieve this, shown how existing attribution losses are unsuitable for CBMs and proposed a new loss that is robust to changes in the learned concepts, and illustrated how the same schema can be used to correct the concepts themselves by interacting with their labels and explanations.  A thorough empirical validation of these ideas is currently underway.

\section*{Acknowledgments}

We are grateful to the anonymous reviewers for helping us to improve the manuscript.
This research has received funding from the European Union's Horizon 2020 FET Proactive project ``WeNet - The Internet of us'', grant agreement No.  823783, and from the ``DELPhi - DiscovEring Life Patterns'' project funded by the MIUR Progetti di Ricerca di Rilevante Interesse Nazionale (PRIN) 2017 -- DD n. 1062 del 31.05.2019.  The research of ST and AP was partially supported by TAILOR, a project funded by EU Horizon 2020 research and innovation programme under GA No 952215.

\bibliography{explanatory-supervision,paper}
\end{document}